\documentclass[]{spie}  

\usepackage{amsmath,amsfonts,amssymb}
\usepackage{graphicx}
\usepackage[colorlinks=true, allcolors=blue]{hyperref}

\usepackage{booktabs}
\usepackage{threeparttable} 

\title{Benchmarking Pretrained Attention-based Models for Real-Time Recognition in Robot-Assisted Esophagectomy}

\author[a]{Ronald L.P.D. de Jong}
\author[a]{Yasmina al Khalil}
\author[a]{Tim J.M. Jaspers}
\author[b]{Romy C. van Jaarsveld}
\author[b]{Gino M. Kuiper}
\author[a]{Yiping Li}
\author[b]{Richard van Hillegersberg}
\author[b]{Jelle P. Ruurda}
\author[a]{Marcel Breeuwer}
\author[a]{Fons van der Sommen}
\affil[a]{Eindhoven University of Technology, Groene Loper 3, 5612 AE Eindhoven, The Netherlands}
\affil[b]{University Medical Center Utrecht, Heidelberglaan 100, 3584 CX Utrecht, The Netherlands}

\authorinfo{Further author information: Ronald L.P.D. de Jong: E-mail: r.l.p.d.d.jong@tue.nl}

\begin{document} 
\maketitle

\begin{abstract}
Esophageal cancer is among the most common types of cancer worldwide. It is traditionally treated using open esophagectomy, but in recent years, robot-assisted minimally invasive esophagectomy (RAMIE) has emerged as a promising alternative. However, robot-assisted surgery can be challenging for novice surgeons, as they often suffer from a loss of spatial orientation. Computer-aided anatomy recognition holds promise for improving surgical navigation, but research in this area remains limited. In this study, we developed a comprehensive dataset for semantic segmentation in RAMIE, featuring the largest collection of vital anatomical structures and surgical instruments to date. Handling this diverse set of classes presents challenges, including class imbalance and the recognition of complex structures such as nerves. This study aims to understand the challenges and limitations of current state-of-the-art algorithms on this novel dataset and problem. Therefore, we benchmarked eight real-time deep learning models using two pretraining datasets. We assessed both traditional and attention-based networks, hypothesizing that attention-based networks better capture global patterns and address challenges such as occlusion caused by blood or other tissues. The benchmark includes our RAMIE dataset and the publicly available CholecSeg8k dataset, enabling a thorough assessment of surgical segmentation tasks. Our findings indicate that pretraining on ADE20k, a dataset for semantic segmentation, is more effective than pretraining on ImageNet. Furthermore, attention-based models outperform traditional convolutional neural networks, with SegNeXt and Mask2Former achieving higher Dice scores, and Mask2Former additionally excelling in average symmetric surface distance.
\end{abstract}

\keywords{Anatomy recognition, cholecystectomy, computer vision, deep learning, esophagectomy, robotics, semantic segmentation, surgery}

\section{Introduction}
\label{sec:Introduction}
Esophageal cancer is the eleventh most common cancer worldwide, and the seventh most common cause of death from cancer~\cite{bray2024global}. Treatment of esophageal cancer generally consists of neoadjuvant chemoradiotherapy followed by esophagectomy~\cite{luketich2000minimally}. Esophagectomy is traditionally performed through open surgery. However, in recent years, robot-assisted minimally invasive esophagectomy (RAMIE) has emerged as an alternative approach. RAMIE minimizes surgical trauma by enabling procedures through small incisions, while the robotic system provides stable movements and tremor suppression. Compared to open surgery, RAMIE leads to fewer complications, shorter hospital stays, and less blood loss during surgery~\cite{na2021robotic, van2019robot, straatman2017minimally}. A drawback of RAMIE is the complexity of the procedure, as is evident from its learning curve of 18-80 cases~\cite{pickering2023learning, van2018learning,  zhang2018learning}. Surgical orientation and identifying crucial anatomical landmarks during RAMIE are particularly challenging for novice surgeons. While the close-up view of the camera on the robot allows for greater visual detail and accurate surgical dissection, it can also lead to a loss of spatial orientation. This is particularly challenging during the thoracic phase of the surgery, where many vital organs are in close proximity to one another. Given the complexity of RAMIE, expert surgeons face a challenge in training novice surgeons. Computer-aided anatomy recognition holds the promise of improving surgical navigation and thereby lowering the learning curve for novice surgeons.

According to a recent systematic review~\cite{den2022computer}, computer-aided anatomy recognition is an emerging research field that is still in its infancy. Many approaches have used deep learning to segment a single organ, while for surgical guidance it is desirable to detect multiple organs or structures. Approaches focusing on multiple organs have mostly been applied to the publicly available CholecSeg8k dataset~\cite{hong2020cholecseg8k}. This dataset consists of 8,080 annotated frames specifically for cholecystectomy procedures, designed to facilitate research in anatomical recognition and semantic segmentation, which involves labeling each pixel in a frame with a specific class label. Conventional convolutional neural networks (CNNs), such as DeepLabv3+~\cite{chen2018encoder}, are commonly used for semantic segmentation in these studies~\cite{silva2022analysis}. Two studies have explored anatomy recognition in RAMIE. Sato et al.~\cite{sato2022real} used DeepLabv3+\cite{chen2018encoder} to recognize the recurrent laryngeal nerve, and den Boer et al.~\cite{den2023deep} utilized a U-Net~\cite{ronneberger2015u} for recognition of the right lung, aorta, vena cava, and azygos vein. Although these studies represent a step towards an effective anatomy recognition system, detecting additional organs is crucial for precise surgical navigation in RAMIE.

In this study, we have created a new dataset for RAMIE, comprising 879 annotated frames from 32 patients across 12 distinct classes, with multiple classes often appearing in a single frame. These classes include four surgical instruments and eight key anatomical structures, making it the most comprehensive set of classes to date. Handling this diverse set of classes presents challenges, including class imbalance and recognizing complex structures such as nerves. This study aims to understand the challenges and limitations of current state-of-the-art algorithms on this novel dataset and problem. To achieve this, we benchmarked eight real-time deep learning models using two pretraining datasets. We evaluated both traditional and attention-based networks, hypothesizing that attention-based networks could capture global patterns and overcome challenges, such as occlusion by blood or other tissues. This benchmark will help us gauge the performance of existing algorithms and identify the limitations that need to be addressed for effective segmentation. The benchmark incorporates our RAMIE dataset alongside the CholecSeg8k dataset, establishing a foundation for surgical anatomy recognition.

\section{Methods}
\label{sec:Methods}
In this section, we outline the methodologies employed in our research. First, Sec.~\ref{sec:RAMIE_dataset} and~\ref{sec:Cholec_dataset} describe the datasets used in our experiments. Sec.~\ref{sec:models_and_pretraining} details the models and pretraining datasets used, while Sec.~\ref{sec:implementation_details} discusses the implementation details. Lastly, Sec.~\ref{sec:evaluation_metrics} outlines the evaluation procedure.

\subsection{RAMIE dataset} \label{sec:RAMIE_dataset}
To create the RAMIE dataset, we acquired surgical recordings of thoracoscopic RAMIE procedures between January 2018 and July 2021 at the University Medical Center Utrecht in The Netherlands.  These recordings included patients who underwent RAMIE for esophageal cancer, with or without neoadjuvant chemoradiotherapy. The procedures were carried out by two expert RAMIE surgeons, each having performed over 200 RAMIE cases. RAMIE typically involves both thoracic and abdominal phases. However, this research focuses exclusively on videos of the thoracic phase, as surgical navigation is more critical during this part of the procedure. The videos were recorded at a frame rate of 25 Hz with a resolution of 960$\times$540 pixels, and have an average duration of two hours. Black borders and the graphical user interface were removed from the recordings to eliminate irrelevant information, resulting in a final resolution of 668$\times$502.

\begin{figure}[t]
\centerline{\includegraphics[width=0.93\textwidth]{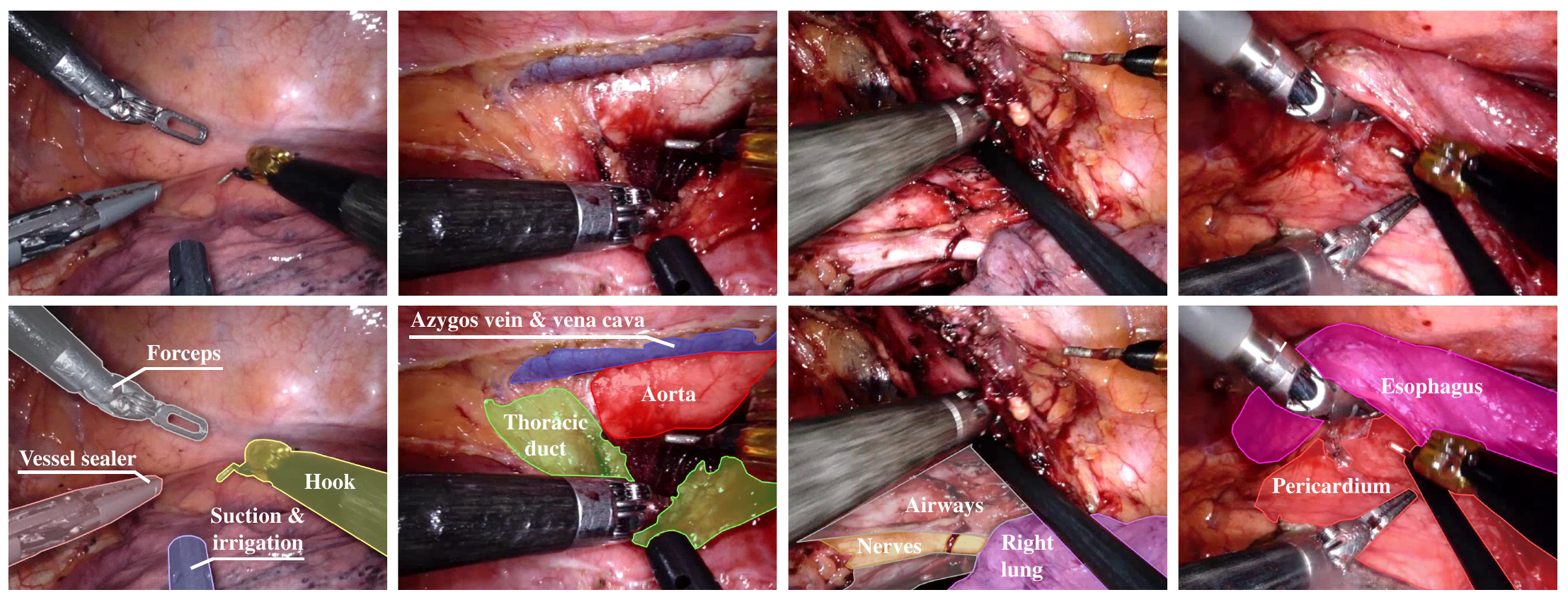}}
\caption{Example frames with corresponding overlays of all distinct classes in the RAMIE dataset. One overlay is shown per class, even when the class is visible in multiple images. \label{fig:class_overview}}
\end{figure}

We randomly sampled 879 frames from the videos of 32 distinct patients. These frames were labeled by four research fellows in the field of surgery and medical imaging under the supervision of an expert surgeon. In total, 12 distinct classes were annotated, including four classes for surgical instruments: forceps, hook, suction \& irrigation, and vessel sealer. The other eight classes are vital anatomical structures appearing during RAMIE: airways, aorta, azygos vein \& vena cava, esophagus, nerves, pericardium, right lung, and thoracic duct. Apart from the 12 distinct classes, a background class was added to create dense semantic labels. Outlines of the different classes are depicted in Fig.~\ref{fig:class_overview}. The airways include the trachea, left main bronchus, and right main bronchus. These structures were depicted as one single class due to their similarity in appearance, as well as difficulties in defining the exact boundary between the trachea and bronchi. For the same reason, the vena cava and azygos vein were combined into a single class, including the subbranches of the azygos vein, technically known as intercostal veins. The nerves class comprises the four most vital nerves during RAMIE: the left and right vagus nerves, along with the left and right recurrent laryngeal nerves. Finally, the pericardium class also includes the pulmonary veins, as this structure is embedded under the same tissue layer.

\begin{figure}[h]
\centerline{\includegraphics[width=0.9\textwidth]{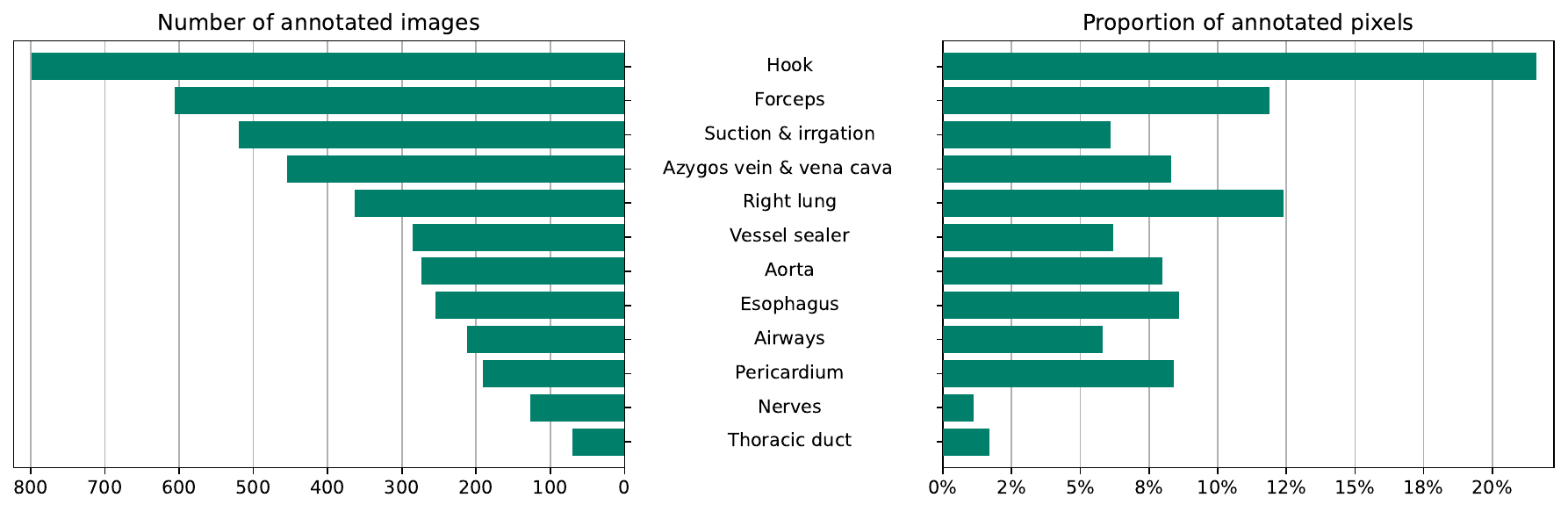}}
\caption{(Left) Number of annotated images per class. (Right) Proportion of annotated pixels per class as a percentage of the total number of annotated pixels.\label{fig:class_occurrences}}
\end{figure}

Fig.~\ref{fig:class_occurrences} shows the number of times each class is present in the dataset. A class imbalance is evident because certain classes only appear during specific phases of the surgery and are thus underrepresented when sampling randomly. Additionally, the classes vary in size, which affects the proportion of annotated pixels per class. For example, the right lung usually appears as a large structure, whereas nerves are usually small.

\subsection{CholecSeg8k dataset} \label{sec:Cholec_dataset}
To allow for a more comprehensive assessment, we additionally used the CholecSeg8k dataset~\cite{hong2020cholecseg8k}, as it has been employed frequently in similar studies~\cite{silva2022analysis, grammatikopoulou2024spatio, zhang2024towards}. This dataset includes sequences of 80 consecutive frames from 101 video fragments, yielding 8,080 semantic segmentation masks. Consistent with previous studies~\cite{grammatikopoulou2024spatio, zhang2024towards}, we excluded low-prevalence classes (blood, cystic duct, hepatic vein, and liver ligament) to ensure a robust analysis.

\subsection{Models and pretraining datasets} \label{sec:models_and_pretraining}
Tab.~\ref{tab:model_overview} provides an overview of eight state-of-the-art models selected for comparison, including their hyperparameters and pretraining type. DeepLabv3, DeepLabv3+, PSPNet, and FPN were chosen for their wide usage in medical imaging and surgical segmentation, while Mask2Former, Segformer, Segmenter, and SegNeXt were selected because they utilize attention mechanisms~\cite{vaswani2017attention}. Attention can help in extracting both local and global features, which is particularly important for segmenting objects of variable size and shape. Attention-based models can also possibly handle occlusion better than traditional CNNs, rendering them suitable for RAMIE.

Due to the limited availability of experts, acquiring annotations for surgical videos tends to be challenging, often resulting in small datasets. Although pretraining offers a solution to mitigate the limitations of such small datasets, large annotated datasets for surgical segmentation are lacking. Therefore, we conducted an evaluation of general computer vision datasets for pretraining. The deep learning models used in this research were pretrained on ImageNet~\cite{deng2009imagenet} and ADE20k~\cite{zhou2017scene}. ImageNet was chosen because of its efficacy across diverse tasks, including medical imaging, whereas ADE20k was chosen due to its specialization in semantic segmentation. ImageNet is a widely used classification dataset in computer vision and contains approximately 1.3 million training images including 1,000 classes. The ADE20k dataset is a well-known dataset for segmentation containing diverse annotations of scenes, objects, and parts of objects. The dataset includes over 20,000 training images with pixel-based labels for 150 distinct classes, such as buildings, cars, and persons. 

\begin{table}[t]
\centering 
\caption{Overview of utilized models, including their encoder, presence of attention (Att.), loss, optimizer, initial learning rate (LR), and used pretraining datasets. \label{tab:model_overview}}%
\small
\begin{tabular*}{\textwidth}{@{\extracolsep\fill}llllllccc@{\extracolsep\fill}}
\toprule
\multicolumn{3}{@{}l}{\textbf{Model details}} & \multicolumn{3}{@{}l}{\textbf{Training details}} & \multicolumn{3}{@{}l}{\textbf{Pretraining datasets}} \\ \cmidrule{1-3} \cmidrule{4-6} \cmidrule{7-9}
\textbf{Model} & \textbf{Encoder} & \textbf{Att.} & \textbf{Loss$^\dagger$}  & \textbf{Optimizer}  & \textbf{LR} & \textbf{Scratch$^\ddagger$} & \textbf{ImageNet} & \textbf{ADE20k}  \\
\midrule
DeepLabv3~\cite{chen2017rethinking}   & ResNet50 & $\times$   & CE        & SGD   & 1e-2 & \checkmark & \checkmark  & \checkmark  \\
DeepLabv3+~\cite{chen2018encoder}     & ResNet50 & $\times$   & CE        & SGD   & 1e-2 & \checkmark & \checkmark  & \checkmark  \\
PSPNet~\cite{zhao2017pyramid}         & ResNet50 & $\times$   & CE        & SGD   & 1e-2 & \checkmark & \checkmark  & \checkmark  \\
FPN~\cite{kirillov2019panoptic}       & ResNet50 & $\times$   & CE        & SGD   & 1e-2 & $\times$   &  $\times$   & \checkmark  \\
Mask2Former~\cite{cheng2022masked}    & ResNet50 & \checkmark & CE+Dice & AdamW & 1e-4 & $\times$   &  $\times$   & \checkmark  \\
Segformer~\cite{xie2021segformer}     & MiT-B0   & \checkmark & CE        & AdamW & 6e-5 & $\times$   &  $\times$   & \checkmark  \\
Segmenter~\cite{strudel2021segmenter} & ViT-S    & \checkmark & CE        & SGD   & 1e-3 & $\times$   &  $\times$   & \checkmark  \\
SegNeXt~\cite{guo2022segnext}         & MSCAN-L  & \checkmark & CE        & AdamW & 6e-5 & $\times$   &  $\times$   & \checkmark  \\
\bottomrule
\end{tabular*}
\begin{tablenotes}
\item \hspace{-0.5cm}$^\dagger$CE is an abbreviation for cross-entropy loss.
$^\ddagger$Scratch indicates the model was trained without pretraining.
\end{tablenotes}
\end{table}

\subsection{Implementation details} \label{sec:implementation_details}
The pretrained ImageNet and ADE20k model weights were obtained from the Segmentation-Models-Pytorch and MMSegmentation packages~\cite{Iakubovskii:2019,mmseg2020}. Subsequently, the models were fine-tuned with fully unfrozen weights using these frameworks. The hyperparameters were kept mostly similar to those found in the original papers of the models, to ensure a consistent evaluation and to avoid biases that could arise from tuning hyperparameters differently among the evaluated models. Since not all weights were available in the used packages, only a subset of models was trained from scratch and pretrained with ImageNet. All models were fine-tuned on the RAMIE and CholecSeg8k datasets. For both datasets, 85\% of the frames were used for training, and 15\% of the frames were used for testing. Within the training set, five-fold cross-validation was applied, where each fold consists of approximately 80\% for training and 20\% for validation. Each set contains data from separate patients to exclude potential biases. Both frames and annotations were resized to 512$\times$512 pixels using bicubic interpolation. The losses used are cross-entropy (CE) or a combination of CE and Dice score. The optimizers used are stochastic gradient descent (SGD) and AdamW~\cite{loshchilov2017decoupled}. Tab.~\ref{tab:model_overview} shows initial learning rates for each model. The learning rate was halved after 10 epochs without validation loss improvement, and early stopping was applied after 25 epochs without improvement. All models were trained on an NVIDIA GeForce RTX 2080 Ti GPU with a batch size of~2. To improve model performance and robustness, augmentations were used, including horizontal flip, vertical flip, blur, brightness, contrast, saturation, scaling, translation, and rotation, all with a probability of 50\%.

\subsection{Evaluation} \label{sec:evaluation_metrics}
We evaluated all models using the Dice score and the average symmetric surface distance (ASSD) \cite{yeghiazaryan2018family}. We selected ASSD over Hausdorff distance because it is less sensitive to outliers. The Dice score ranges from 0 to 1, whereas ASSD is measured in pixels on a 512$\times$512 frame. A high Dice score and a low ASSD are preferable. Metrics were calculated on a per-image basis, and we assessed statistical significance using a Wilcoxon signed-rank test. For visual evaluation, the predictions were scaled back to the original image size using bicubic interpolation.

\section{Results}
\label{sec:Results}
This section presents the benchmark results in three parts. Sec.~\ref{sec:pretraining_evaluation} compares the performance of the ImageNet and ADE20k pretraining datasets using a subset of models. Sec.~\ref{sec:quantitative_model_evaluation} offers a quantitative analysis of both attention-based and non-attention-based models utilizing the best pretraining dataset. Finally, Sec.~\ref{sec:qualitative_model_evaluation} provides a qualitative evaluation on a selected subset of models.

\subsection{Quantitative pretraining evaluation} \label{sec:pretraining_evaluation}
Tab.~\ref{tab:results_pretraining} presents performance metrics for DeepLabv3, DeepLabv3+, and PSPNet pretrained on various pretraining datasets and fine-tuned on the RAMIE and CholecSeg8k datasets. On both datasets, the models pretrained on ADE20k significantly outperform those pretrained on ImageNet or without pretraining. This could be explained by the fact that ADE20k is specific to segmentation, which closely aligns with the fine-tuning task. Since ADE20k pretraining yields the best performance, it is used for the remaining experiments.

\begin{table*}[h]
\centering
\caption{Performance metrics for DeepLabv3, DeepLabv3+, and PSPNet pretrained using various pretraining datasets and fine-tuned on the RAMIE and CholecSeg8k datasets.   \label{tab:results_pretraining}}
\small
\begin{tabular*}{\textwidth}{@{\extracolsep\fill}llcccc@{}}
\toprule
&&\multicolumn{2}{@{}l}{\textbf{RAMIE dataset}} & \multicolumn{2}{@{}l}{\textbf{CholecSeg8k dataset}} \\\cmidrule{3-4}\cmidrule{5-6}
\textbf{Model} & \textbf{Pretraining} & \textbf{Dice score}  & \textbf{ASSD [pixels]}   & \textbf{Dice score}  & \textbf{ASSD [pixels]}  \\
\midrule
DeepLabv3      & No pretraining  & 0.43 $\pm$ 0.14 & 38 $\pm$ 26 & 0.53 $\pm$ 0.09 & 44 $\pm$ 16 \\
               & ImageNet        & 0.52 $\pm$ 0.16 & 29 $\pm$ 21 & 0.58 $\pm$ 0.11 & 34 $\pm$ 14 \\
               & ADE20k          & \textbf{\phantom{$^{*}$}0.56 $\pm$ 0.14$^{*}$} & \textbf{\phantom{$^{*}$}23 $\pm$ 18$^{*}$} & \textbf{\phantom{$^{*}$}0.61 $\pm$ 0.11$^{*}$} & \textbf{\phantom{$^{*}$}30 $\pm$ 13$^{*}$} \\
\midrule
DeepLabv3+     & No pretraining  & 0.39 $\pm$ 0.14 & 44 $\pm$ 29 & 0.50 $\pm$ 0.09 & 42 $\pm$ 15 \\
               & ImageNet        & 0.51 $\pm$ 0.16 & 29 $\pm$ 23 & 0.56 $\pm$ 0.11 & 36 $\pm$ 13 \\
               & ADE20k          & \textbf{\phantom{$^{*}$}0.57 $\pm$ 0.15$^{*}$} & \textbf{\phantom{$^{*}$}24 $\pm$ 21$^{*}$} & \textbf{\phantom{$^{*}$}0.60 $\pm$ 0.11$^{*}$} & \textbf{\phantom{$^{*}$}32 $\pm$ 13$^{*}$} \\
\midrule
PSPNet         & No pretraining  & 0.38 $\pm$ 0.13 & 45 $\pm$ 30 & 0.52 $\pm$ 0.10 & 38 $\pm$ 15  \\
               & ImageNet        & 0.53 $\pm$ 0.17 & 31 $\pm$ 25 & 0.57 $\pm$ 0.11 & 34 $\pm$ 10 \\
               & ADE20k          & \textbf{\phantom{$^{*}$}0.57 $\pm$ 0.16$^{*}$} & \textbf{\phantom{$^{*}$}27 $\pm$ 22$^{*}$} & \textbf{\phantom{$^{*}$}0.60 $\pm$ 0.11$^{*}$} & \textbf{\phantom{$^{*}$}30 $\pm$ 13$^{*}$}  \\
\bottomrule
\end{tabular*}
\begin{tablenotes}[flushleft] 
\item \hspace{-0.15cm} $^{*}p<0.05$ using a Wilcoxon signed-rank test. Results are reported as mean $\pm$ standard deviation computed across all images in the test set, with the best metric scores indicated in bold.
\end{tablenotes}
\end{table*}

\subsection{Quantitative model evaluation} \label{sec:quantitative_model_evaluation}

\begin{table}[b]
\footnotesize
\caption{FPS and performance metrics for various models pretrained on the ADE20k dataset. FPS was calculated on an NVIDIA GeForce RTX 2080 Ti GPU and rounded to the nearest integer. \label{tab:results_models}}
\small
\begin{tabular*}{\textwidth}{@{\extracolsep\fill}lccccc@{}}
\toprule
&  &\multicolumn{2}{@{}l}{\textbf{RAMIE dataset}} & \multicolumn{2}{@{}l}{\textbf{CholecSeg8k dataset}} \\\cmidrule{3-4}\cmidrule{5-6}
\textbf{Model} & \textbf{FPS} & \textbf{Dice score}  & \textbf{ASSD [pixels]}  & \textbf{Dice score}  & \textbf{ASSD [pixels]}  \\
\midrule
DeepLabv3       & 87  & 0.56 $\pm$ 0.14 & 23 $\pm$ 18  & 0.61 $\pm$ 0.11 & 30 $\pm$ 13  \\
Deeplabv3+      & \textbf{100} & 0.57 $\pm$ 0.15 & 24 $\pm$ 21  & 0.60 $\pm$ 0.11 & 32 $\pm$ 13  \\
PSPNet          & 96  & 0.57 $\pm$ 0.16 & 27 $\pm$ 22  & 0.60 $\pm$ 0.11 & 30 $\pm$ 13  \\
FPN             & 87  & 0.50 $\pm$ 0.16 & 33 $\pm$ 27 & 0.54 $\pm$ 0.10 & 34 $\pm$ 13 \\
Mask2Former     & 19  &  \textbf{\phantom{$^{*}$}0.71 $\pm$ 0.16$^{*}$} & \textbf{\phantom{$^{*}$}11 $\pm$ 10$^{*}$}  & \textbf{\phantom{$^{*}$}0.73 $\pm$ 0.11$^{*}$} & \textbf{\phantom{$^{*}$}20 $\pm$ 11$^{*}$} \\
Segformer       & 69  & 0.64 $\pm$ 0.15 & 17 $\pm$ 13  & 0.67 $\pm$ 0.08 & 25 $\pm$ 11  \\
Segmenter       & 30  & 0.68 $\pm$ 0.15 & 15 $\pm$ 11  & 0.69 $\pm$ 0.10 & 23 $\pm$ 12 \\
SegNeXt         & 25  &  \textbf{\phantom{$^{*}$}0.71 $\pm$ 0.14$^{*}$} & 14 $\pm$ 12 & \textbf{\phantom{$^{*}$}0.73 $\pm$ 0.08$^{*}$} & 22 $\pm$ 10  \\
\bottomrule
\end{tabular*}
\begin{tablenotes}[flushleft] 
\item \hspace{-0.15cm} $^{*}p<0.05$ using a Wilcoxon signed-rank test. Results are reported as mean $\pm$ standard deviation computed across all images in the test set, with the best metric scores indicated in bold.
\end{tablenotes}
\end{table}

\noindent Tab.~\ref{tab:results_models} displays the frames per second (FPS) and performance metrics for various models pretrained on ADE20k. Among the evaluated models, the traditional CNNs (DeepLabv3, DeepLabv3+, PSPNet, and FPN) achieve the highest FPS, partly because they do not rely on complex attention mechanisms. However, attention-based networks (Segformer, Mask2Former, Segmenter, and SegNeXt) excel in terms of segmentation quality. In particular, SegNeXt and Mask2Former significantly outperform all other models in terms of Dice Score. Although there is no significant difference between the Dice scores of these two models, Mask2Former achieves a significantly lower ASSD compared to all other models. Despite 19-25 FPS being on the lower end for smooth human perception, it is generally acceptable in surgical settings where rapid movements are rare.

\begin{figure}[t]
  \centering
  \begin{minipage}{0.56\textwidth}
    \centering
    \includegraphics[width=\textwidth, trim=0pt 0.25cm 0pt 0.2cm, clip]{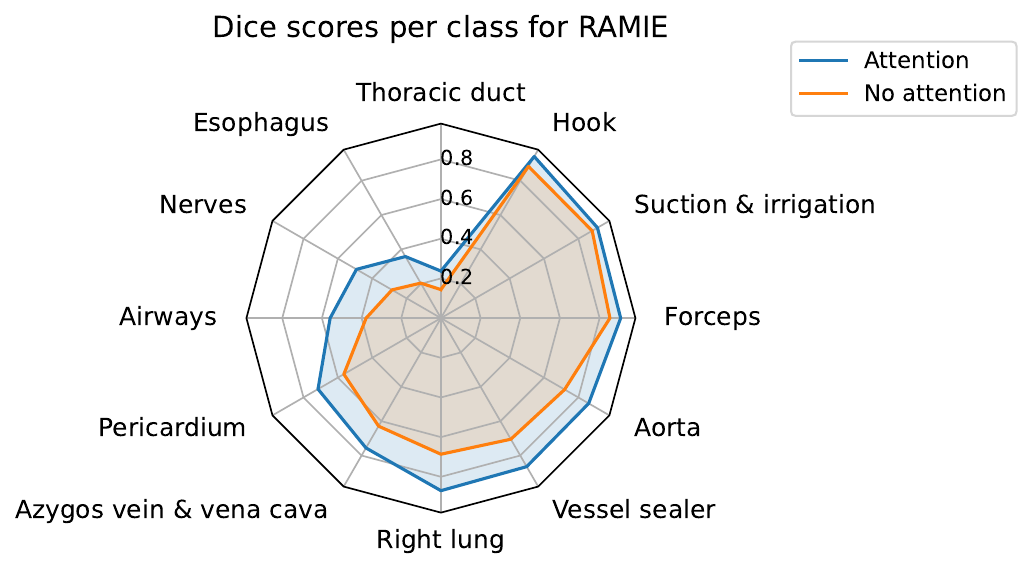}
  \end{minipage}
  \begin{minipage}{0.43\textwidth}
    \centering
    \includegraphics[width=\textwidth, trim=0pt 0.2cm 0pt 0.2cm, clip]{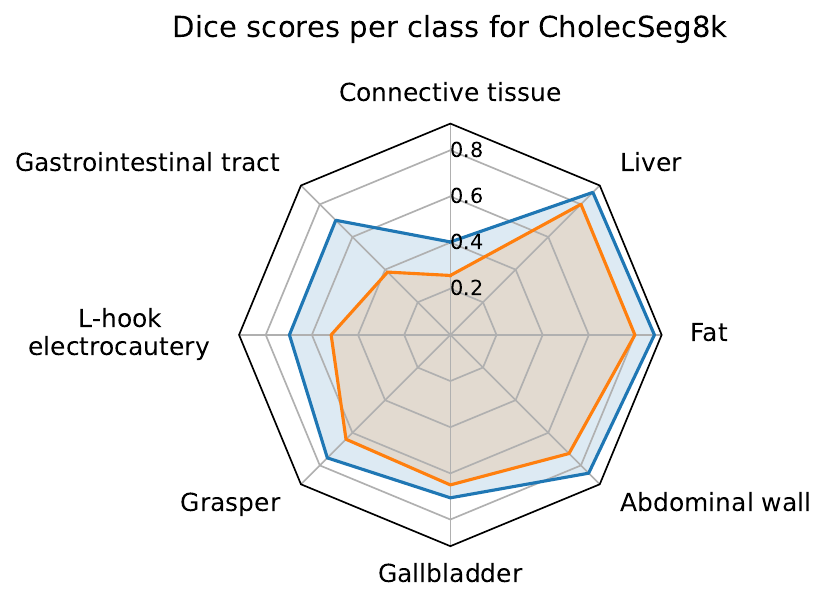}
  \end{minipage}
  \vspace{0.1cm}
  \caption{Dice scores per class for RAMIE (left) and CholecSeg8k (right), averaged across models with attention (Mask2Former, Segformer, Segmenter, SegNeXt) and without attention (DeepLabv3, DeepLabv3+, PSPNet, FPN).}
  \label{fig:radar}
\end{figure}

Fig. \ref{fig:radar} shows Dice scores per class for RAMIE and CholecSeg8k, averaged across all models with and without attention. It can be observed that the attention-based models achieve higher Dice scores on average across all classes. For RAMIE, it is evident that there is variation in performance per class, with lower performance for underrepresented and smaller anatomical structures such as nerves and the thoracic duct. 

\subsection{Qualitative model evaluation} \label{sec:qualitative_model_evaluation}

\begin{figure}[b]
\centerline{\includegraphics[width=0.93\textwidth]{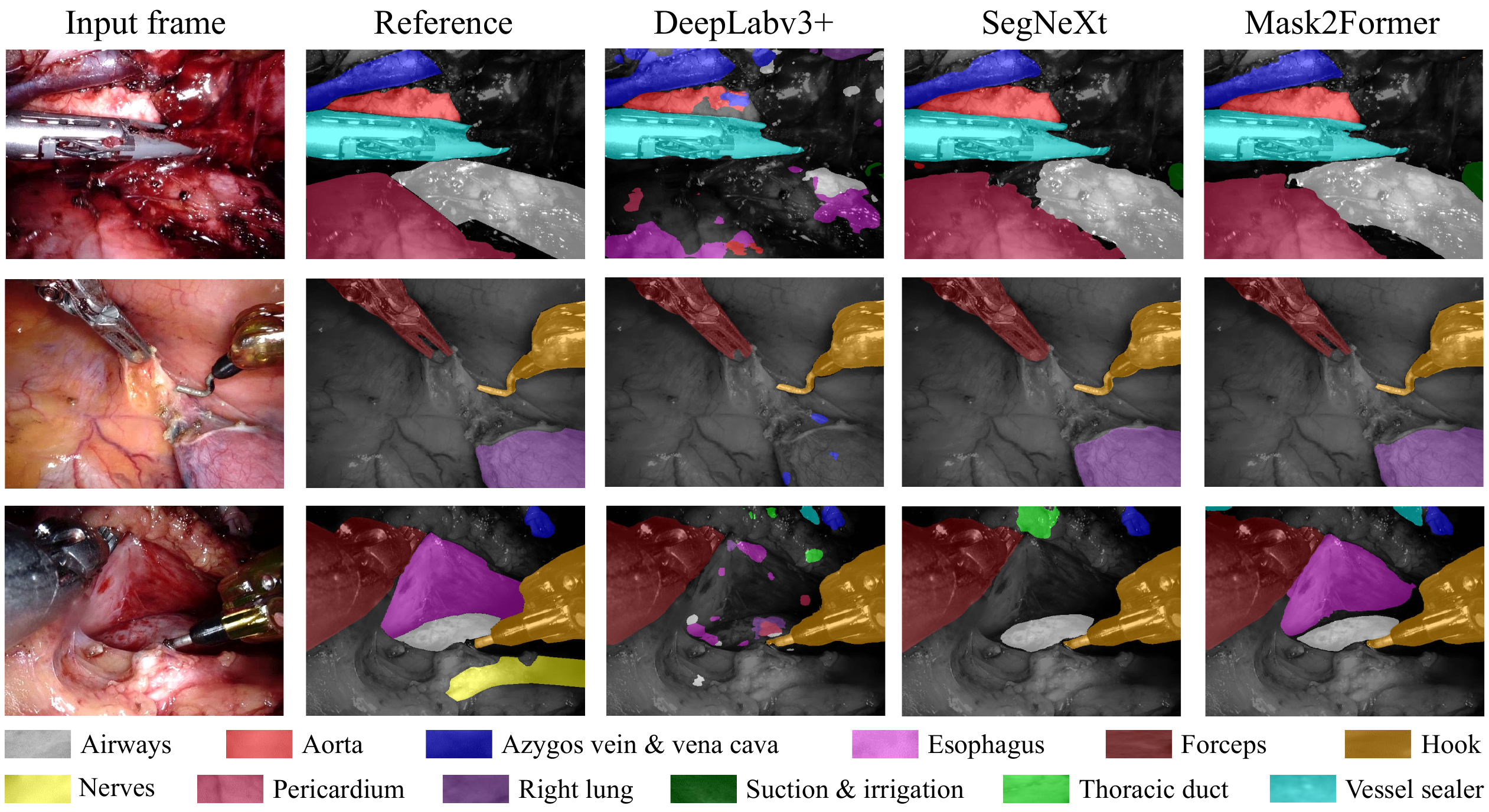}}
\caption{Visualization of input frames, reference annotations, and predictions on the RAMIE dataset using DeepLabv3+, SegNeXt, and Mask2Former, each pretrained on ADE20k.}
\label{fig:visual_results_RAMIE}
\end{figure}

Fig.~\ref{fig:visual_results_RAMIE} presents model predictions on the RAMIE dataset using DeepLabv3+, SegNeXt, and Mask2Former. We have selected SegNeXt and Mask2Former for visual evaluation, as they demonstrated the highest performance in our quantitative model evaluation in Sec.~\ref{sec:quantitative_model_evaluation}. Additionally, we included DeepLabv3+ to facilitate a visual comparison with a non-attention-based model, providing a broader perspective on the impact of attention mechanisms on segmentation outcomes. In Fig. \ref{fig:visual_results_RAMIE}, it can be observed that the surgical tools are well segmented by all models. In the top row, DeepLabv3+ provides partial segmentation of the azygos vein and aorta. However, its predictions for the airways and pericardium are compromised due to blood obscuring these structures. In contrast, SegNeXt and Mask2Former accurately predict these structures, likely because attention-based models focus on global representations rather than local textures. In the second row, DeepLabv3+ fails to segment the lung, which is partly outside the field of view. SegNeXt and Mask2Former, however, successfully segment this structure. The bottom row illustrates a scenario where all models encounter challenges, especially in detecting the nerve, likely due to its partial embedding in tissue and the underrepresentation of this class in the dataset.

\begin{figure}[h]
\centerline{\includegraphics[width=0.93\textwidth]{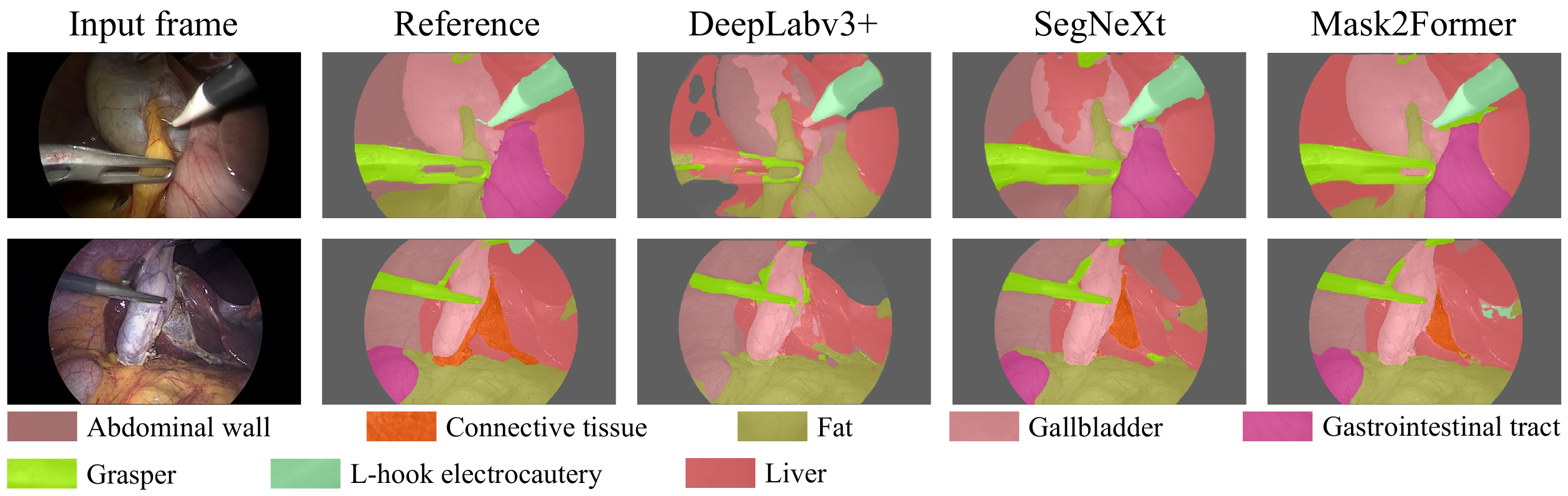}}
\caption{Visualization of input frames, reference annotations, and predictions on the CholecSeg8k dataset using DeepLabv3+, SegNeXt, and Mask2Former, each pretrained on ADE20k.}
\label{fig:visual_results_Cholec}
\end{figure}

Figure \ref{fig:visual_results_Cholec} illustrates model predictions for the CholecSeg8k dataset. Both rows demonstrate improved detection of the gastrointestinal tract by attention-based models. Additionally, the bottom row shows that connective tissue is more accurately identified by these models. Finally, attention-based models also achieve a more precise delineation of surgical tools.

\section{Discussion}
\label{sec:Discussion}
In this study, we made an initial attempt at constructing a large semantic segmentation dataset for RAMIE procedures, with the eventual goal of making it freely accessible. To the best of our knowledge, it is the most comprehensive dataset for semantic segmentation in RAMIE to date, encompassing a wide range of anatomical structures and surgical instruments. We evaluated multiple pretraining datasets and models using both this unique dataset and the publicly available CholecSeg8k dataset to identify the most effective approach for addressing the challenges in surgical segmentation, as well as to uncover the limitations of current methods.

From the pretraining evaluation on the RAMIE and CholecSeg8k datasets, it can be concluded that pretraining leads to better segmentation performance, most likely due to the small sizes of the surgical datasets. Additionally, models pretrained on ADE20k perform best, which could be explained by the fact that this is a semantic segmentation dataset, and therefore closer to the final application than ImageNet. Given these findings, it is recommended to investigate pretraining on more segmentation datasets in future research. 

Based on the comparison of models on the RAMIE and CholecSeg8k datasets, it can be concluded that the ones that do use attention outperform those that do not. An explanation for this could be that attention can capture long-range dependencies between pixels or regions in an image, which is more difficult to achieve with traditional CNNs. Furthermore, attention allows for focusing on relevant objects even in the presence of occlusions or clutter in the frames. Notably, attention-based models exhibit a lower FPS compared to convolutional models, highlighting a tradeoff between high segmentation quality and low inference time. Nevertheless, attention-based models remain capable of operating near or in real-time, rendering them suitable for surgical anatomy recognition tasks. Mask2Former and SegNeXt especially show superior performance and are therefore recommended for future research. Mask2Former additionally excels at precise boundary delineation, as indicated by its low ASSD. This may be attributed to its use of masked attention, which enables focused processing of specific areas of interest.

The performance gain of attention-based models over traditional CNNs is particularly notable in classes that are underrepresented, occluded by blood, or partially obstructed by instruments. On RAMIE, all models achieve strong segmentation results for surgical instruments, owing to their frequent occurrence and distinct appearance in the dataset. Furthermore, the models score well on the aorta, azygos vein \& vena cava, and right lung, which usually have high contrast. The model achieves lower scores for the airways and pericardium, which are less prevalent in the dataset. Segmentation accuracy for the esophagus is limited, despite its frequent appearance in the procedure. One potential explanation is the variation in the visual appearance of the esophagus throughout the surgical procedure. The low segmentation performance for the nerves and thoracic duct may be attributed to their rare appearance in the dataset and their smaller size compared to other anatomical structures. Additionally, in our dataset, the thoracic duct was annotated along with the surrounding fat, making it difficult to distinguish from regular fat tissue. Nerves, on the other hand, can be difficult to detect since they are often embedded in connective tissue. However, the nerves and thoracic duct are among the most important anatomical structures for surgeons. Therefore, it is vital to include more data from these specific classes in future research.

A limitation of this study is that we have primarily focused on the presence of attention mechanisms rather than the specific types of attention used. Additionally, other factors that affect performance, such as model size, were not considered. In future research, we will address these aspects with a more extensive analysis. An additional limitation is the exclusive use of general computer vision pretraining datasets. To explore the benefits of in-domain pretraining, we plan to extend the benchmark by incorporating pretraining on surgical data through self-supervised learning, as proposed in our prior study\cite{10.1007/978-3-031-73748-0_5}. Finally, future work should assess the relevance of anatomy recognition models to surgical practice by investigating the significance of included anatomical structures and ensuring evaluation metrics align with clinical needs.

\section{Conclusion}
\label{sec:Conclusion}
In this study, we have developed a comprehensive dataset for semantic segmentation in RAMIE, including a wide range of anatomical structures and surgical instruments. Through the evaluation of various pretraining datasets and models on both our RAMIE dataset and the publicly available CholecSeg8k dataset, we have identified key pretraining and model features, as well as highlighted significant challenges in surgical segmentation. With this work, we hope to facilitate future studies aimed at enhancing surgical navigation, potentially reducing the learning curve for novice surgeons. 

\section{Acknowledgments}
\label{sec:Acknowledgment}
This research was funded by Stichting Hanarth Fonds, study number: 2022-13. It is part of the INTRA-SURGE (INTelligent computeR-Aided Surgical gUidance for Robot-assisted surGEry) project aimed at advancing the future of surgery.

\bibliography{report} 
\bibliographystyle{spiebib} 

\end{document}